\documentclass[letterpaper, 10 pt, conference]{ieeeconf}
\IEEEoverridecommandlockouts  %
\usepackage{amsmath}
\usepackage{amssymb}
\usepackage{mathtools}
\usepackage{graphicx}
\usepackage{adjustbox}
\let\labelindent\relax
\usepackage{enumitem}
\usepackage{booktabs}
\usepackage{bm}
\usepackage{acro}
\usepackage{makecell}
\usepackage{array}
\usepackage{multirow}
\usepackage{subcaption}
\usepackage{xcolor}
\usepackage{pgfplots}
\usepackage{amsmath}
\usepackage{cuted}
\usepackage{fontawesome5}
\usepackage[most]{tcolorbox}
\usepackage{wrapfig}
\usepackage{pgfplots}
\pgfplotsset{compat=1.18}

\makeatletter\let\NAT@parse\undefined\makeatother
\usepackage[bookmarks=true]{hyperref}
\usepackage[capitalize]{cleveref}

\DeclareAcronym{DDP}{
  short = DDP,
  long  = differential dynamics programming
}
\DeclareAcronym{MPCC}{
  short = MPCC,
  long  = mathematical program with complementarity constraints,
  long-plural-form = mathematical programs with complementarity constraints,
}
\DeclareAcronym{NLP}{
  short = NLP,
  long  = nonlinear program
}
\DeclareAcronym{AL}{
  short = AL,
  long  = augmented Lagrangian
}
\DeclareAcronym{CVF}{
  short = CVF,
  long  = conditional value function
}

\newcommand{\diag}{\mathop{\mathrm{diag}}}

\AtBeginDocument{%
  \thinmuskip=3mu
  \medmuskip=4mu
  \thickmuskip=5mu
  \binoppenalty=10000
  \relpenalty=10000
  \linepenalty=1000
}

\newcolumntype{C}[1]{>{\centering\arraybackslash}p{#1}}

\newcommand{\best}[1]{{\textcolor{blue!70!black}{#1}}}
\newcommand{\second}[1]{{\textcolor{green!50!black}{#1}}}
\newcommand{\rev}[1]{{\color{black!80!black}#1}}
\newcommand{\revtwo}[1]{{\color{black!85!black}#1}}

\pgfplotsset{compat=1.18}

\newtcbox{\linkpill}{
  on line,
  box align=base,
  colback=gray!10,
  colframe=gray!10,
  arc=1.5mm,
  boxrule=0pt,
  left=2pt,
  right=2pt,
  top=1pt,
  bottom=1pt
}

\title{\LARGE\bf 
Lifted Multiple Shooting for Parallel-in-Time \\ Contact-Rich Trajectory Optimization
}

\author{Wei-Chen Li and Glen Chou
\thanks{The authors are with the Georgia Institute of Technology, Atlanta, GA 30308, USA. \texttt{\{\href{mailto:wli777@gatech.edu}{wli777},\href{mailto:chou@gatech.edu}{chou}\}@gatech.edu}}%
}

\begin{document}

\maketitle
\thispagestyle{empty}
\pagestyle{empty}

\begin{abstract}

Planning through contact requires accounting for the coupled dynamics of a robot and its environment, including inertial effects, contact forces, and friction. 
However, the non-smoothness and multimodality of contact dynamics and contact geometry makes gradient-based optimization challenging. 
In this work, we propose lifted multiple shooting, a trajectory optimization transcription that can be solved directly using nonlinear program solvers. 
By introducing contact impulses as optimization variables, our formulation allows the optimizer to satisfy the state-transition constraints using friction-cone-feasible impulses while separately enforcing their consistency with the contact model.
Our formulation supports a broad class of contact constitutive relations, with sensitivities computed through implicit differentiation.
To enable efficient planning over long horizons and for high-dimensional systems, we further introduce IP-PAS, a parallel-in-time primal-dual interior-point solver that achieves $\mathcal{O}(\log T)$ parallel computational depth. 
We evaluate the proposed methods on six fully dynamic contact-rich planning tasks, including nonprehensile pushing and in-hand manipulation, without contact-guiding costs. 
Lifted multiple shooting improves success rates by up to 81.5 percentage points over standard multiple shooting, while IP-PAS achieves solution quality comparable to serial methods with up to a 3.1$\times$ speedup and scales to horizons of 40{,}000 timesteps.

\begin{center}
\vspace{-0.5\baselineskip}
\linkpill{\href{https://trustworthyrobotics.github.io/contact-planning-NLP-site/}{\textcolor{blue!50!black}{\faGlobe\ Website}}}
\quad
\linkpill{\href{https://trustworthyrobotics.github.io/contact-planning-NLP-site/code}{\textcolor{blue!50!black}{\faGithub\ Code}}}
\end{center}

\end{abstract}

\section{Introduction}

Trajectory optimization through contact has largely followed two approaches: contact smoothing (\cref{tbl:prior-work}, columns 1 and 2) and contact-implicit optimization (\cref{tbl:prior-work}, column 3).
Contact smoothing methods regularize contact dynamics to provide useful gradients across contact transitions, enabling gradient-based optimization without explicitly reasoning about contact modes.
A related perspective arises in reinforcement learning, where policy-gradient methods that optimize expected returns over a distribution of dynamics can be interpreted as performing stochastic smoothing of the optimization landscape \cite{pang2023global}.
In contrast, contact-implicit methods introduce contact forces as optimization variables and enforce contact laws through complementarity constraints.

\begin{figure}[t]
    \centering
    \includegraphics[width=\columnwidth]{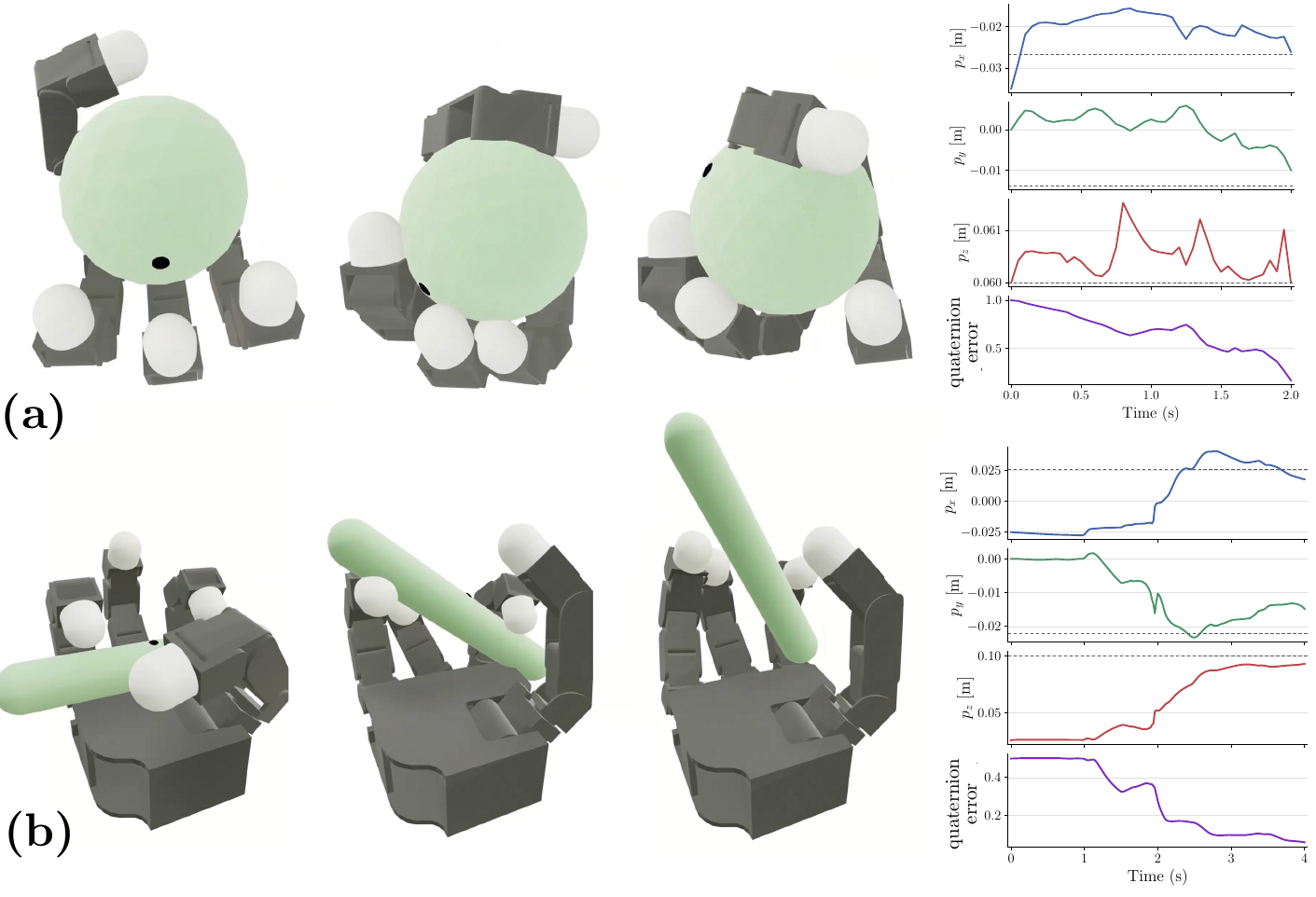}\vspace{-5pt}
    \caption{\small\looseness-1\textbf{In-hand tasks}. We show our solver, IP-PAS, with lifted multiple shooting generating a trajectory for \textbf{(a)} an Allegro hand rotating a sphere \revtwo{$180^\circ$ over a 40-timestep planning horizon} and \textbf{(b)} grasping and lifting a pen from its palm and rotating it such that it rests vertically against the fingers \revtwo{over a 200-timestep horizon}. The rightmost column shows absolute 3D position over time (\revtwo{black dotted lines indicate the goal positions}) and quaternion goal error. \vspace{-18pt}}
    \label{fig:inhand-tasks}
\end{figure}

Despite their apparent differences, we show in \cref{sec:preliminaries} that these approaches can be understood as different transcriptions of the same underlying bilevel trajectory optimization problem.
While contact smoothing methods yield smooth \acp{NLP}, contact-implicit formulations lead to \acp{MPCC}.
However, \acp{MPCC} violate standard constraint qualifications at all feasible points, preventing their direct treatment as smooth \acp{NLP} with standard convergence guarantees.
Consequently, existing contact-implicit methods typically either reformulate the \ac{MPCC} as a smooth \ac{NLP} through complementarity relaxation or penalization, or develop specialized algorithms for handling complementarity constraints.

This distinction raises a question: Can we formulate trajectory optimization through contact as a standard \ac{NLP} while retaining the advantages of explicitly optimizing contact forces?
Conventional shooting formulations eliminate contact forces from the optimization variables, making trajectory improvement dependent on the sensitivities of the contact-resolution problem.
Conversely, contact-implicit formulations expose contact forces to the optimizer but introduce complementarity constraints that complicate numerical optimization.
We investigate whether an alternative transcription can combine the advantages of both approaches without requiring specialized \ac{MPCC} solvers.

Our contributions are as follows:
\begin{itemize}[leftmargin=*]
\item We propose a lifted multiple shooting transcription that exposes contact impulses as optimization variables while avoiding complementarity constraints in the outer NLP.
\item We formulate contact resolution as a convex optimization problem for a broad class of contact constitutive relations, and use implicit differentiation to compute the sensitivities.
\item We derive a stagewise primal-dual interior-point KKT system and reformulate its serial backward and forward recursions as associative scans, enabling parallel-in-time trajectory optimization.
\item We compare trajectory optimization transcriptions and solvers across pushing and in-hand manipulation tasks, evaluating optimization reliability and computational scalability without task-specific warm starts and contact-guiding costs.
\end{itemize}

\section{Related Work and Preliminaries}\label{sec:preliminaries}

Multibody dynamics with generalized coordinates $q$ and generalized velocity $v$ can be written in discrete time as
\begin{subequations}\small
\begin{align}
    & M(q)\, (v^+ - v) + \delta t\, c(q, v) = \delta t\, \tau(q^+, v^+, u) + J(q)^\top \gamma,  \label{eq:manipulator-equation} \\
    & q^+ = q \oplus \delta t\, v^+,  \label{eq:velocity-integration}
\end{align}
\end{subequations}
where $\delta t$ is the timestep, $x = (q,v)$ and $x^+ = (q^+,v^+)$ denote the current and next states, respectively. Here, $M(q)$ is the mass matrix, $c(q,v)$ collects Coriolis, centrifugal, and gravitational forces, and $\tau(q^+,v^+,u)$ represents stiffness, damping, and actuation forces. The matrix $J(q)$ is the contact Jacobian, and $\gamma$ denotes the contact impulse. The operator $\oplus$ denotes the appropriate integration operation for both vector and quaternion coordinates. 
Computing the contact Jacobian $J(q)$ requires a closest-point query/collision detection between geometries, which also provides a signed distance $\phi(q)$.
Dynamics of deformable bodies can be described using the same equations \cite{li2026convex}.
When $\tau$ is linearized with respect to $v$, the velocity update can be written explicitly as
\begin{equation}  \label{eq:velocity-dynamics}
    v^+ = v^* + M_e^{-1} J(q)^\top \gamma, 
\end{equation}
where
\begin{equation}  \label{eq:velocity-dynamics-def}
\begin{aligned}
    M_e &:= M(q) - \delta t\, \bigl( \tfrac{\partial \tau}{\partial v} (q,v,u) + \delta t\, \tfrac{\partial \tau}{\partial q} (q,v,u) \bigr), \\
    v^* &:= v + \delta t\, M_e^{-1} \bigl( \tau(q,v,u) - c(q,v) \bigr).
\end{aligned}    
\end{equation}
Together, \eqref{eq:velocity-dynamics} and \eqref{eq:velocity-integration} define the mapping
\begin{equation}
x^+ = f_\text{integrate}(x,u,\gamma).
\end{equation}

The contact impulse $\gamma$ is not known a priori. Instead, it is determined by solving a contact-resolution problem whose parameters depend on the current state and control $(x,u)$. Thus, a numerical contact solver induces a mapping
\begin{equation}
\gamma = f_\text{contact}(x,u),
\end{equation}
and the overall dynamics can be viewed as the composition
\begin{equation} \label{eq:dynamics}
x^+ = f_\text{step}(x, u) := f_\text{integrate}\bigl( x, u, f_\text{contact}(x, u) \bigr).
\end{equation}

\subsection{Contact Models}
We next review contact models commonly used in planning through contact.

\subsubsection{Signorini condition, Coulomb friction, and Maximum dissipation principle}
Under the rigid contact assumption, where rigid bodies cannot interpenetrate, the contact-resolution problem can be written as
\begin{equation}
\small
\begin{aligned}
    \text{find} \quad & \gamma  \\
    \text{s.t.} \quad 
        & M_e (v^+ - v^*) = J(q)^\top \gamma,  \\
        & \textrm{v}_c := \begin{bmatrix} \mathrm{v}_n \\ \mathrm{v}_\mathrm{tan} \end{bmatrix} = J(q)\, v^+,  \\
        & 
            \begin{cases}
                \gamma = 0                                                      & \text{if } \textrm{v}_n + \tfrac{\phi(q)}{\delta t} > 0,  \\
                \gamma \in \mathcal{F}_\mu                                      & \text{if } \textrm{v}_c + [\begin{matrix} \tfrac{\phi(q)}{\delta t} & 0 & 0 \end{matrix}]^\top = 0,  \\
                \gamma_{\mathrm{tan}} = -\mu \gamma_n \tfrac{\mathrm{v}_{\mathrm{tan}}}{\|{\mathrm{v}_{\mathrm{tan}}}\|} & \text{otherwise},
            \end{cases}
\end{aligned}
\end{equation}
where $\mathcal{F}_\mu$ is the friction cone with friction coefficient $\mu$.

These conditions define a non-convex, potentially set-valued contact-resolution problem \cite{RN860}. Thus, the physical model does not uniquely determine $\gamma$ as a function of $(x,u)$. A numerical solver selects one solution from this set, thereby inducing the mapping \(f_{\mathrm{contact}}\); neighboring \((x,u)\) values can therefore yield drastically different solver-selected impulses and unreliable gradients. %

\subsubsection{Log-barrier}
By employing Anitescu's relaxation \cite{RN765}, the contact-resolution problem can be written as
\begin{subequations}
\begin{align}
    \text{find} \quad & \gamma  \\
    \text{s.t.} \quad 
        & M_e (v^+ - v^*) = J(q)^\top \gamma,  \label{eq:log-barrier-stationarity}  \\
        & \mathcal{F}_\mu \ni \gamma \perp J(q)\, v^+ + [\begin{matrix} \tfrac{\phi(q)}{\delta t} & 0 & 0 \end{matrix}]^\top \in \mathcal{F}_\mu^*,  \label{eq:log-barrier-complementarity}
\end{align}
\end{subequations}
where $\mathcal{F}_\mu^*$ denotes the dual cone of $\mathcal{F}_\mu$.
{The} stationarity condition \eqref{eq:log-barrier-stationarity} and complementarity condition \eqref{eq:log-barrier-complementarity} are precisely the KKT conditions of a convex optimization problem with a log-barrier \cite{pang2023global}. However, this contact model still assumes rigid contact and has the hydroplaning artifact, where tangential motion generates a normal force \cite{suh2026dexterous}.

\subsubsection{Complementarity-free}
Another contact model \cite{jin2025complementarity} computes the contact impulse in closed form as
\begin{equation}
    \gamma = \delta t\, k \cdot \mathrm{relu}\bigl( -\phi(q \oplus \delta t\, v^*) \bigr),
\end{equation}
where $k$ is the contact stiffness. 
In other words, the model treats contact as compliant by evaluating the gap distance at a configuration predicted from $v^*$ and setting the contact impulse proportional to the resulting penetration distance.

Because $\gamma$ is computed independently of $v^+$ and $q^+$, rather than being coupled to the discrete dynamics through an implicit constitutive relation, the method corresponds to an explicit integration scheme. Consequently, it is limited to small contact stiffnesses to maintain numerical stability.

\subsection{Transcriptions for Bilevel Trajectory Optimization}
We next review common transcriptions for solving trajectory optimization problems with the dynamics in \eqref{eq:dynamics}.

\subsubsection{Single shooting}
Single shooting transcribes the trajectory optimization problem as
\begin{equation}
    \min_{u_{0:T-1}} \ \textstyle\sum_{t=0}^{T-1} \ell(x_t, u_t) + \ell_T(x_T),
\end{equation}
where the dynamics $x_{t+1} = f_\text{step}(x_t, u_t)$ are satisfied by construction by rolling out the system from an initial condition $\hat{x}_0$ under the control sequence $u_{0:T-1}$. 
Thus, only the controls are optimization variables. 
Single shooting problems are often solved via \ac{DDP}.

\subsubsection{Multiple shooting}
Multiple shooting instead transcribes the trajectory optimization problem as
\begin{equation}
\begin{aligned}
    \min_{x_{0:T}, u_{0:T-1}} \
        & \textstyle\sum_{t=0}^{T-1} \ell(x_t, u_t) + \ell_T(x_T) \\
    \text{s.t.} \hspace{1.8em} 
        & x_0 = \hat{x}_0, \\
        & x_{t+1} = f_\text{step}(x_t, u_t), \quad t=0,\dots,T-1.   
\end{aligned}    
\end{equation}
Unlike single shooting, multiple shooting treats the states as optimization variables and enforces the dynamics as equality constraints, allowing dynamically infeasible initial trajectories and often improving conditioning. 

It is worth noting that trajectory optimization through contact can be viewed as bilevel optimization problems, with an outer trajectory optimization and an inner contact-resolution problem. In both single and multiple shooting, the contact solver must provide the required sensitivities $\partial f/\partial x$ and $\partial f/\partial u$ to the outer optimization problem so that it can compute a descent direction. In bilevel optimization, such methods are called \emph{descent methods} \cite{sinha2018bilevel}.

\subsubsection{Contact implicit}
Contact-implicit trajectory optimization incorporates the KKT conditions of the inner contact-resolution problem directly into the outer trajectory optimization problem. For example, using the log-barrier contact model, the resulting problem takes the form
\begin{equation}
\hspace{-8pt}\begin{aligned}
    \min_{\substack{x_{0:T}, \\ u_{0:T-1},\\\gamma_{0:T-1}}} \
        & \textstyle\sum_{t=0}^{T-1} \ell(x_t, u_t) + \ell_T(x_T) \hspace*{-3em} \\
    \text{s.t.} \hspace{1.em}  
        & x_0 = \hat{x}_0, \\
        & x_{t+1} = f_\text{integrate}(x_t, u_t, \gamma_t), && t=0,\dots,T-1,  \\
        & \mathcal{F}_\mu \ni \gamma_t \perp J(q_t)\, v_{t+1} + [\begin{matrix} \tfrac{\phi(q_t)}{\delta t} & 0 & 0 \end{matrix}]^\top \in \mathcal{F}_\mu^*,  \hspace*{-50em}  \\
        &                                  && t=0,\dots,T-1.
\end{aligned}
\end{equation}
Rather than solving the contact-resolution problem separately at each timestep, this formulation incorporates the contact impulses and feasibility conditions directly into the trajectory optimization, resulting in an \ac{MPCC}. In bilevel optimization, this is a \emph{KKT single-level reduction} \cite{sinha2018bilevel}. 

Beyond the difficulty of solving an \ac{MPCC}, the problem can grow substantially when $J(q)$ and $\phi(q)$ lack closed-form expressions for geometries such as polyhedrons. In these cases, contact forces are introduced for each candidate face, together with complementarity constraints enforcing that at most one face is active \cite{li2025on}.

In summary, existing approaches to trajectory optimization through contact, whether termed contact smoothing or contact implicit, solve the same underlying bilevel trajectory optimization problem using different mathematical transcriptions. \cref{tbl:prior-work} categorizes prior work. 

This raises the question of whether a better transcription can preserve the bilevel structure without reducing it to a single-level problem with complementarity constraints, while improving optimization reliability to avoid poor local minima from naive warm starts. In the following, we will outline our novel lifted multiple shooting transcription (Sec. \ref{sec:transcription}), and our parallel-in-time \ac{NLP} solver for efficiently solving the transcription (Sec. \ref{sec:solver}).

\begin{table}[t]
    \centering
    \caption{Trajectory optimization through contact methods.}
    \label{tbl:prior-work}
    \setlength{\tabcolsep}{1.5pt}
    \begin{tabular}{cccc}
        \toprule
        Contact model & Single shooting & Multiple shooting & Contact implicit \\
        \midrule
        \makecell{Signorini condition\\\& Coulomb friction} 
            & \cite{howell2022traj, kim2025contact} 
            & \cite{caleach2024fast}
            & \makecell{\cite{manchester2019contact, aydinoglu2024consensus, bui2025push},\\\cite{zhang2025simultaneous, li2025on, li2026impact}} \\
        \midrule
        \makecell{Log-barrier} 
            & \makecell{\cite{suh2026dexterous, li2026certified, jiang2025robust},\\\cite{zhang2026simultaneous}}
            & -
            & \cite{aydinoglu2024consensus, jin2025complementarity} \\
        \midrule
        \makecell{Complementarity-free} 
            & -
            & \cite{jin2025complementarity, kurtz2026inverse}
            & - \\
        \bottomrule
    \end{tabular}\vspace{-12pt}
\end{table}

\section{Problem Transcription and Contact Solver}\label{sec:transcription}

We transcribe the trajectory optimization problem as
\begin{subequations} \label{eq:lms}
\small
\begin{align}
    \hspace{-1em}
    \min_{\substack{x_{0:T}, u_{0:T-1},\\\gamma_{0:T-1}}} \
        & \textstyle\sum_{t=0}^{T-1} \ell(x_t, u_t) + \ell_T(x_T) \hspace{-3em} \\
    \text{s.t.} \hspace{1.7em}  
        & x_0 = \hat{x}_0, \\
        & x_{t+1} = f_\text{integrate}(x_t, u_t, \gamma_t), && t=0,\dots,T-1,  \label{eq:lms-integrate} \\
        & \gamma_t = f_\text{contact}(x_t, u_t),          && t=0,\dots,T-1,  \label{eq:lms-contact} \\
        & \gamma_t \in \mathcal{F}_\mu,                 && t=0,\dots,T-1.  \label{eq:lms-friction-cone}
\end{align}
\end{subequations}
We call this transcription \emph{lifted multiple shooting}.
In contrast to standard multiple shooting, contact impulses are introduced as explicit decision variables. This allows the optimizer to satisfy the state-transition constraints \eqref{eq:lms-integrate} using any friction-cone-feasible contact impulses
\eqref{eq:lms-friction-cone}, while separately enforcing their consistency through the contact model \eqref{eq:lms-contact}.
In contrast to contact implicit, the bilevel structure is preserved and we don't need to solve an \ac{MPCC}.

We next describe how $\gamma = f_\text{contact}(x, u)$ is implicitly defined. At each timestep, the contact impulse is obtained by solving a convex contact-resolution problem. The contact-resolution problem is
\begin{equation} \label{eq:contact-resolution}
    \min_{v^+} \quad \textstyle\frac{1}{2} (v^+ - v^*)^\top M_e (v^+ - v^*) + \ell_c(\mathrm{v}_c),
\end{equation}
where $M_e$ and $v^*$ are defined in \eqref{eq:velocity-dynamics-def}, $\mathrm{v}_c := J(q)\, v^+$ is the contact velocity and $\ell_c$ is a convex regularization term that will be defined later.

The first-order optimality condition of \eqref{eq:contact-resolution} is
\begin{equation} \label{eq:contact-optimality}
    \textstyle M_e\, (v^+ - v^*) + J(q)^\top\, \frac{\partial \ell_c(\mathrm{v}_c)}{\partial \mathrm{v}_c} = 0.
\end{equation}
Comparing \eqref{eq:contact-optimality} with the discrete-time dynamics in \eqref{eq:velocity-dynamics}, we obtain the relationship between the contact impulse and \revtwo{the} regularization term $\ell_c$:
\begin{equation} \label{eq:impulse-cost-relation}
    \textstyle \gamma = -\frac{\partial \ell_c(\mathrm{v}_c)}{\partial \mathrm{v}_c}.
\end{equation}
\rev{Thus, for a constitutive relation $\gamma = \gamma(\mathrm{v}_c)$, $\ell_c$ is defined by $\nabla_{\mathrm{v}_c}\ell_c(\mathrm{v}_c)=-\gamma(\mathrm{v}_c)$.} 
\rev{Since $M_e$ is positive definite and $\ell_c$ is convex whenever $-\partial \gamma(\mathrm{v}_c)/\partial \mathrm{v}_c \succeq 0$, the objective in \eqref{eq:contact-resolution} is strongly convex in $v^+$.} Therefore, the contact-resolution problem admits a unique optimal velocity $v^+$. The corresponding contact impulse is obtained from \eqref{eq:impulse-cost-relation}, which implicitly defines the mapping $\gamma = f_\text{contact}(x,u)$.

Although the contact-resolution problem is formulated in terms of the convex regularization term $\ell_c$, its explicit evaluation is not required by our solver. We use Newton's method to solve the first-order optimality condition
\begin{equation}
    r(v^+;x,u) := M_e\, (v^ + -v^*) - J(q)^\top\, \gamma(v_c) = 0,
\end{equation}
A Newton direction is obtained by solving
\begin{equation}
    \textstyle H\, \Delta v^+ = -r,
    \qquad
    H := M_e - J(q)^\top \frac{\partial\gamma}{\partial v_c} J(q) .
\label{eq:contact-hessian}
\end{equation}
An exact line search determines the step size by finding the zero of the directional derivative, which can likewise be evaluated using only the constitutive relation $\gamma(v_c)$ and its derivative \cite{castro2025icf}. Thus, only the first- and second-order derivatives of the scalar potential $\ell_c$ are required by the solver.

The same Hessian enables efficient implicit differentiation of the contact solution. Since the optimal velocity satisfies $r(v^+;x,u)=0$, differentiating this equation with respect to any parameter $p\in\{x,u\}$ gives
\begin{equation} \label{eq:contact-implicit-diff}
    \textstyle H\, \frac{\partial v^+}{\partial p} = -\frac{\partial r}{\partial p}.
\end{equation}
The right-hand side is obtained by forward-mode automatic differentiation of the residual function, while the sensitivities are computed by solving a linear system using the already factorized Hessian $H$. Thus, implicit differentiation avoids differentiating through the Newton iterations and makes computing the contact sensitivities inexpensive relative to solving the contact problem itself.

For the contact constitutive relation, we use the Hunt-Crossley model \cite{hunt1975coefficient} for the normal contact impulse. For a contact with \revtwo{current signed distance $\phi_0:=\phi(q)$} and relative normal velocity $\mathrm{v}_n$, the impulse is given by
\begin{equation} \label{eq:hunt-crossley}
    \gamma_n = \delta t\, k \cdot
        \operatorname{relu}_{\epsilon}(-\phi_0 - \delta t\, \mathrm{v}_n) \cdot
        \operatorname{relu}(1 - d \mathrm{v}_n),
\end{equation}
where $k$ and $d$ are the contact stiffness and Hunt-Crossley damping, respectively.
Here, $\operatorname{relu}_{\epsilon}$ denotes a smooth approximation of the rectified linear unit.
We use
\begin{minipage}[c]{0.68\linewidth}
    \begin{equation*}
        \operatorname{relu}_{\epsilon}(-\phi)
        =
        \begin{cases}
            \textstyle\frac{\epsilon^2}{4(\epsilon+\phi)},
                & \phi \geq -\epsilon/2, \\[3pt]
            -\phi, & \phi < -\epsilon/2.
        \end{cases}
    \end{equation*}
\end{minipage}
\hfill
\begin{minipage}[c]{0.28\linewidth}
    \centering
    \begin{tikzpicture}
\begin{axis}[
    height=3.3cm,
    axis lines=middle,
    axis equal image,
    xlabel={$\phi$},
    ylabel={},
    xmin=-1.5, xmax=1.5,
    ymin=0, ymax=1.6,
    xtick={0},
    ytick=\empty,
    tick label style={font=\scriptsize},
    label style={font=\scriptsize},
    clip=false,
]
    \addplot[black, dashed, semithick, domain=-1.5:1.5,
             samples=100] {max(-x,0)};
    
    \addplot[blue!30!black, thick, domain=-1.5:1.5,
             samples=200]
        {x < -0.25 ? -x : 0.25/(4*(0.5+x))};
\end{axis}
\end{tikzpicture}

\end{minipage}
Smoothing is introduced as a necessary numerical compromise. In the lifted multiple shooting transcription \eqref{eq:lms}, the equality constraint $\gamma_t=f_{\text{contact}}(x_t,u_t)$
must be corrected whenever it is violated. For a nonsmooth contact model, however, $f_{\text{contact}}$ may have zero gradient, providing insufficient first-order information for the optimizer to reduce the constraint violation. We observe that the lifted multiple shooting transcription requires substantially less smoothing than single- or multiple-shooting transcriptions: in our experiments, it operates reliably with a smoothing parameter approximately $1000\times$ smaller. This substantially reduces the deviation of the smoothed contact model from the original nonsmooth dynamics.

\section{Parallel-in-Time Solver}\label{sec:solver}

We introduce an IP-PAS (interior-point parallel associative scan) solver utilizing temporal parallelization to solve \eqref{eq:lms}.

Define the augmented stage control $w_t := (u_t, \gamma_t)$. 
Then \eqref{eq:lms} has the generic form
\begin{subequations}\label{eq:ocp}
\small
\begin{align}
    \min_{x_{0:T},w_{0:T-1}}\quad
        & \textstyle\sum_{t=0}^{T-1} \ell(x_t,w_t)+\ell_T(x_T)  \hspace{-3em} \\
    \mathrm{s.t.}\quad
        & x_0 = \hat{x}_0,                                   \label{eq:ocp-initial-eq}\\
        & x_{t+1} = f(x_t, w_t),  && t=0,\dots,T-1,           \label{eq:ocp-dyn}\\
        & h(x_t, w_t) = 0,        && t=0,\dots,T-1,           \label{eq:ocp-eq}\\
        & g(x_t, w_t) \geq 0,     && t=0,\dots,T-1,           \label{eq:ocp-ineq}\\
        & g_T(x_T) \geq 0.                                    \label{eq:ocp-terminal-ineq}
\end{align}
\end{subequations}

The Lagrangian of \eqref{eq:ocp} is given by
\begin{equation}
\begin{aligned}
    \mathcal{L}
    =& \textstyle\sum_{t=0}^{T-1} \ell(x_t, w_t) + \ell_T(x_T) + \lambda_0^\top (\hat{x}_0 - x_0) \\
    &+ \textstyle\sum_{t=0}^{T-1} \lambda_{t+1}^\top \bigl( f(x_t, w_t) - x_{t+1} \bigr) \\
    &+ \textstyle\sum_{t=0}^{T-1} \beta_t^\top h(x_t, w_t)  \\
    &- \textstyle\sum_{t=0}^{T-1} z_t^\top g(x_t, w_t) - z_T^\top g_T(x_T) ,
\end{aligned}
\end{equation}
and has optimality conditions
\begin{subequations} \label{eq:ocp-optimality}
\begin{align}
    & \nabla_{x_t} \mathcal{L} = 0,       && t=0,\dots,T,    \\ 
    & \nabla_{w_t} \mathcal{L} = 0,       && t=0,\dots,T-1,  \\
    & \nabla_{\lambda_t} \mathcal{L} = 0, && t=0,\dots,T,    \\
    & \nabla_{\beta_t} \mathcal{L} = 0,   && t=0,\dots,T-1,  \\
    & g(x_t, w_t) - s_t = 0,              && t=0,\dots,T-1,  \\
    & g_T(x_T) - s_T = 0,                                    \\
    & \diag(z_t)\, s_t = \kappa 1,        && t=0,\dots,T,    \\  
    & z_t \geq 0,\ s_t \geq 0,            && t=0,\dots,T,  
\end{align}
\end{subequations}
with the interior-point barrier parameter $\kappa \to 0$.

\subsection{Stagewise Optimal Control Structure}
We derive the primal-dual interior-point KKT step for \eqref{eq:ocp-optimality}, which admits stagewise structure.
\subsubsection{Terminal Stage}
Define the terminal stage residual as
\begin{subequations}
\begin{align}
    r_T^x &:= \nabla \ell_T(x_T) - \nabla g_T(x_T)^\top\, z_T - \lambda_T,  \\
    r_T^g &:= g_T(x_T) - s_T,  \\
    r_T^c &:= \diag(z_T)\, s_T - \kappa 1.
\end{align}
\end{subequations}
The KKT step satisfies
\begin{subequations}
\begin{alignat}{2}
    & Q_T\, \delta x_T - G_T^\top\, \delta z_T - \delta \lambda_T &&= -r_T^x,  \label{eq:terminal-kkt-x} \\
    & G_T\, \delta x_T - \delta s_T &&= -r_T^g,                                \label{eq:terminal-kkt-g} \\
    & Z_T\, \delta s_T + S_T\, \delta z_T &&= -r_T^c,                          \label{eq:terminal-kkt-c}
\end{alignat}
\end{subequations}
where 
\begin{align*}
    Q_T &:= \nabla_{x_T}^2 \mathcal{L} = \nabla_{x_T}^2 \ell_T(x_T) - z_T^\top\, \nabla_{x_T}^2 g_T(x_T) ,  \\
    G_T &:= \nabla_{x_T} g_T(x_T),  \\
    Z_T &:= \diag(z_T),\quad S_T := \diag(s_T).
\end{align*}
Equations \eqref{eq:terminal-kkt-g} and \eqref{eq:terminal-kkt-c} give
\begin{subequations} \label{eq:terminal-delta-s-z}
\begin{align}
    \delta s_T &= G_T\, \delta x_T + r_T^g, \\
    \delta z_T &= -S_T^{-1} Z_T G_T\, \delta x_T - S_T^{-1} (Z_T\, r_T^g + r_T^c).  \label{eq:terminal-delta-z}
\end{align}
\end{subequations}
Substituting \eqref{eq:terminal-delta-z} into \eqref{eq:terminal-kkt-x} gives the relation between the terminal state and costate
\begin{equation} \label{eq:terminal-state-costate}
    \delta \lambda_T = P_T\, \delta x_T + p_T,
\end{equation}
where
\begin{align*}
    P_T &= Q_T + G_T^\top S_T^{-1} Z_T G_T,  \\
    p_T &= r_T^x + G_T^\top S_T^{-1} (Z_T\, r_T^g + r_T^c).
\end{align*}

\subsubsection{Intermediate Stage}
Define the intermediate stage ($t=0,\dots,T-1$) residual as
\begin{subequations}
\begin{align}
    r_t^x &:= \nabla_x \ell_t + \nabla_x f_t^\top\, \lambda_{t+1} + \nabla_x h_t^\top\, \beta_t - \nabla_x g_t^\top\, z_t - \lambda_t,  \\
    r_t^w &:= \nabla_w \ell_t + \nabla_w f_t^\top\, \lambda_{t+1} + \nabla_w h_t^\top\, \beta_t - \nabla_w g_t^\top\, z_t,  \\
    r_t^f &:= f(x_t, w_t) - x_{t+1},  \\
    r_t^h &:= h(x_t, w_t),  \\
    r_t^g &:= g(x_t, w_t) - s_t,  \\
    r_t^c &:= \diag(z_t)\, s_t - \kappa 1.
\end{align}
\end{subequations}
\newpage\noindent
\mbox{The KKT step satisfies}
\begin{strip}
\resizebox{\textwidth}{!}{%
\begin{minipage}[c]{0.65\textwidth}
\begin{subequations}
\begin{alignat}{2}
    & Q_t\, \delta x_t + N_t\, \delta w_t + A_t^\top\, \delta \lambda_{t+1} + (H_t^x)^\top\, \delta \beta_t - (G_t^x)^\top\, \delta z_t - \delta \lambda_t &&= -r_t^x,  \label{eq:stage-kkt-x} \\
    & N_t^\top\, \delta x_t + R_t\, \delta w_t + B_t^\top\, \delta \lambda_{t+1} + (H_t^w)^\top\, \delta \beta_t - (G_t^w)^\top\, \delta z_t &&= -r_t^w,                \label{eq:stage-kkt-u} \\
    & A_t\, \delta x_t + B_t\, \delta w_t - \delta x_{t+1} &&= -r_t^f,                                                                                           \label{eq:stage-kkt-f} \\
    & H_t^x\, \delta x_t + H_t^w\, \delta w_t &&= -r_t^h,                                                                                                        \label{eq:stage-kkt-h} \\
    & G_t^x\, \delta x_t + G_t^w\, \delta w_t - \delta s_t &&= -r_t^g,                                                                                           \label{eq:stage-kkt-g} \\
    & Z_t\, \delta s_t + S_t\, \delta z_t &&= -r_t^c,                                                                                                            \label{eq:stage-kkt-c}
\end{alignat}
\end{subequations}
\end{minipage}
\hspace{1em}
\begin{minipage}[c]{0.28\textwidth}  \vspace{0.8em}
\begin{equation*}
    \begin{bmatrix} Q_t & N_t \\ N_t^\top & R_t \end{bmatrix} := \nabla^2_{(x_t, w_t)} \mathcal{L},
\end{equation*} \vspace{-1.5em}
\begin{alignat*}{2}
    A_t &:= \nabla_{x_t} f(x_t, w_t),\quad && B_t := \nabla_{w_t} f(x_t, w_t),   \\
    H_t^x &:= \nabla_{x_t} h(x_t, w_t),    && H_t^w := \nabla_{w_t} h(x_t, w_t),  \\
    G_t^x &:= \nabla_{x_t} g(x_t, w_t),    && G_t^w := \nabla_{w_t} g(x_t, w_t),  \\
    Z_t &:= \diag(z_t),                    && S_t := \diag(s_t).
\end{alignat*}
\end{minipage}
}
\end{strip}
\noindent
Equations \eqref{eq:stage-kkt-g} and \eqref{eq:stage-kkt-c} give
\begin{subequations} \label{eq:stage-delta-s-z}
\begin{align}
    \delta s_t &= G_t^x\, \delta x_t + G_t^w\, \delta w_t + r_t^g, \\
    \delta z_t &= -S_t^{-1} Z_t (G_t^x\, \delta x_t + G_t^w\, \delta w_t) - \xi_t,  \label{eq:stage-delta-z}
\end{align}
\end{subequations}
where $\xi_t = S_t^{-1} (Z_t r_t^g + r_t^c)$.
Substituting \eqref{eq:stage-delta-z} into \eqref{eq:stage-kkt-x} and \eqref{eq:stage-kkt-u} gives
\begin{subequations}
\begin{alignat}{2}
    & \hat{Q}_t\, \delta x_t + \hat{N}_t\, \delta w_t + A_t^\top\, \delta \lambda_{t+1} + (H_t^x)^\top\, \delta \beta_t - \delta \lambda_t &&= -\hat{r}_t^x,  \label{eq:stage-kkt-xhat} \\
    & \hat{N}_t^\top\, \delta x_t + \hat{R}_t\, \delta w_t + B_t^\top\, \delta \lambda_{t+1} + (H_t^w)^\top\, \delta \beta_t &&= -\hat{r}_t^w,     \label{eq:stage-kkt-uhat}                                                           
\end{alignat}
\end{subequations}
where
\begin{align*}
    \begin{bmatrix} \hat{Q}_t & \hat{N}_t \\ \hat{N}_t^\top & \hat{R}_t \end{bmatrix}
    &= \begin{bmatrix} Q_t & N_t \\ N_t^\top & R_t \end{bmatrix}
    + \begin{bmatrix} G_t^x & G_t^w \end{bmatrix}^\top S_t^{-1} Z_t \begin{bmatrix} G_t^x & G_t^w \end{bmatrix}, \\
    \begin{bmatrix} \hat{r}_t^x \\ \hat{r}_t^w \end{bmatrix}
    &= \begin{bmatrix} r_t^x \\ r_t^w \end{bmatrix}
    + \begin{bmatrix} G_t^x & G_t^w \end{bmatrix}^\top \xi_t.
\end{align*}
Equations \eqref{eq:stage-kkt-xhat}, \eqref{eq:stage-kkt-uhat}, \eqref{eq:stage-kkt-f}, and \eqref{eq:stage-kkt-h} give a system of linear equations over $\delta x_t$, $\delta w_t$, $\delta \lambda_{t+1}$, $\delta \beta_t$, and $\delta \lambda_t$. To eliminate $\delta w_t$ and $\delta \beta_t$, we write \eqref{eq:stage-kkt-uhat} and \eqref{eq:stage-kkt-h} as
\begin{equation}  \label{eq:stage-delta-u-beta}
    M_t \begin{bmatrix} \delta w_t \\ \delta \beta_t \end{bmatrix}
    + K_t\, \delta x_t
    + L_t\, \delta \lambda_{t+1}
    + e_t = 0,
\end{equation}
where
\begin{equation*}
    M_t \!=\! \begin{bmatrix} \hat{R}_t & (H_t^w)^\top \\ H_t^w & 0 \end{bmatrix}\!,\
    K_t \!=\! \begin{bmatrix} \hat{N}_t^\top \\ H_t^x \end{bmatrix}\!,\
    L_t \!=\! \begin{bmatrix} B_t^\top \\ 0 \end{bmatrix}\!,\
    e_t \!=\! \begin{bmatrix} \hat{r}_t^w \\ r_t^h \end{bmatrix},
\end{equation*}
and $M_t$ is primal-dual regularized if required to have the correct inertia.
Substituting \eqref{eq:stage-delta-u-beta} into \eqref{eq:stage-kkt-xhat} and \eqref{eq:stage-kkt-f} yields
\begin{subequations} \label{eq:stage-state-costate}
\begin{align}
    \delta \lambda_t &= P_t\, \delta x_t + p_t + \bar{A}_t^\top\, \delta \lambda_{t+1},  \\
    \delta x_{t+1} &= \bar{A}_t\, \delta x_t + b_t - C_t\, \delta \lambda_{t+1},
\end{align}
\end{subequations}
where
\begin{align*}
    P_t &= \hat{Q}_t - K_t^\top M_t^{-1} K_t,   \\
    p_t &= \hat{r}_t^x - K_t^\top M_t^{-1} e_t,   \\
    \bar{A}_t &= A_t - L_t^\top M_t^{-1} K_t,  \\
    b_t &= r_t^f - L_t^\top M_t^{-1} e_t,   \\
    C_t &= L_t^\top M_t^{-1} L_t.
\end{align*}

\subsubsection{Backward and Forward Pass}
Suppose for all $t=0,\cdots,T$, state and costate satisfy the generalized value function relation
\begin{equation}  \label{eq:all-state-costate}
    \delta \lambda_t = P_t^V\, \delta x_t + p_t^V.
\end{equation}
Using \eqref{eq:terminal-state-costate}, the backward pass starts from
\begin{equation}
    P_T^V = P_T, \quad p_T^V = p_T,
\end{equation}
and using \eqref{eq:stage-state-costate} we obtain the backward recursion
\begin{subequations} \label{eq:backward-recursion}
\begin{align}
    P_t^V &= P_t + \bar{A}_t^\top \bigl( I + P_{t+1}^V C_t \bigr)^{-1} P_{t+1}^V \bar{A}_t,  \\
    p_t^V &= p_t + \bar{A}_t^\top \bigl( I + P_{t+1}^V C_t \bigr)^{-1} \bigl( P_{t+1}^V\, b_t + p_{t+1}^V \bigr).
\end{align}
\end{subequations}
Finally, the forward pass starts from
\begin{equation}
    \delta x_0 = \hat{x}_0 - x_0,
\end{equation}
with the forward recursion
\begin{equation} \label{eq:forward-recursion}
    \delta x_{t+1} = \bigl( I + C_t P_{t+1}^V \bigr)^{-1}\, \bigl( \bar{A}_t\, \delta x_t + b_t - C_t\, p_{t+1}^V \bigr).
\end{equation}

\subsection{Temporal Parallel Backward and Forward Pass}
Rather than the serial recursions \eqref{eq:backward-recursion} and \eqref{eq:forward-recursion} which have $\mathcal{O}(T)$ computational depth, the backward and forward pass can be reformulated to yield $\mathcal{O}(\log T)$ computational depth and $\mathcal{O}(T)$ total work, resulting in a parallel runtime of $\mathcal{O}(T/N_{\mathrm{core}}+\log T)$ on $N_{\mathrm{core}}$ compute cores \cite{blelloch1990prefixsum}.

\subsubsection{Parallel Backward Pass}
We define the 5-tuple
\begin{equation*}
    \mathcal{E}_{i \to j} := (P_{ij}, p_{ij}, \bar{A}_{ij}, b_{ij}, C_{ij})
\end{equation*}
as describing the boundary relation between $i \to j$:
\begin{subequations} \label{eq:boundary-state-costate}
\begin{align}
    \delta \lambda_i &= P_{ij}\, \delta x_i + p_{ij} + \bar{A}_{ij}^\top\, \delta \lambda_j,  \\
    \delta x_j &= \bar{A}_{ij}\, \delta x_i + b_{ij} - C_{ij}\, \delta \lambda_j.
\end{align}
\end{subequations}
Equation \eqref{eq:stage-state-costate} yields
\begin{equation}
    \mathcal{E}_{t \to t+1} = (P_t, p_t, \bar{A}_t, b_t, C_t),
\end{equation}
for $t=0,\dots,T-1$. Similarly, \eqref{eq:terminal-state-costate} yields
\begin{equation}
    \mathcal{E}_{T \to T+1} = (P_T, p_T, 0, 0, 0).
\end{equation}
Here, $T+1$ is a dummy time index that facilitates the encoding \eqref{eq:terminal-state-costate} as \eqref{eq:boundary-state-costate}.
For the backward pass, we eventually want $\mathcal{E}_{t \to T+1}$ for all $t=0,\cdots,T$, which gives us $P_t^V$ and $p_t^V$ according to
\begin{equation}
    \mathcal{E}_{t \to T+1} = (P_t^V, p_t^V, 0, 0, 0).
\end{equation}

For a composition
\begin{equation}
    \mathcal{E}_{i \to j} \circ \mathcal{E}_{j \to k} = \mathcal{E}_{i \to k},
\end{equation}
it can be shown, by eliminating $\delta x_j$, $\delta \lambda_j$ and comparing coefficients, that the composition rule is
\begin{subequations}
\begin{align}
    P_{ik} &=P_{ij} +\bar A_{ij}^{\top}(I+P_{jk}C_{ij})^{-1}P_{jk}\bar A_{ij},\\
    p_{ik} &=p_{ij} +\bar A_{ij}^{\top}(I+P_{jk}C_{ij})^{-1} (p_{jk}+P_{jk}\, b_{ij}),\\
    \bar A_{ik} &=\bar A_{jk}(I+C_{ij}P_{jk})^{-1}\bar A_{ij},\\
    b_{ik} &=b_{jk} +\bar A_{jk}(I+C_{ij}P_{jk})^{-1} (b_{ij}-C_{ij}\, p_{jk}),\\
    C_{ik} &=C_{jk} +\bar A_{jk}(I+C_{ij}P_{jk})^{-1} C_{ij}\bar A_{jk}^{\top}.
\end{align}
\end{subequations}
This composition rule is associative and is algebraically identical to that derived for single shooting optimal control in \cite{sarkka2023temporal}. 
However, our derivation is based on the primal-dual KKT system of multiple shooting, including the proposed lifted multiple shooting.

The quantities of interest $\mathcal{E}_{t \to T+1} = (P_t^V, p_t^V, 0, 0, 0)$ for $t=0,\dots,T$ can be computed from the suffix compositions
\begin{equation}
    \mathcal{E}_{t \to T+1} = \mathcal{E}_{t \to t+1} \circ \mathcal{E}_{t+1 \to t+2} \circ \cdots \circ \mathcal{E}_{T \to T+1},
\end{equation}
which when using associative scans has a computational depth of $\mathcal{O}(\log T)$.

\subsubsection{Parallel Forward Pass}
We define the 2-tuple
\begin{equation}
    \mathcal{D}_{i \to j} := (D_{ij}, d_{ij})
\end{equation}
as describing the forward relation between $i \to j$:
\begin{equation}
    \delta x_j = D_{ij}\, \delta x_i + d_{ij}.
\end{equation}
Equation \eqref{eq:forward-recursion} yields
\begin{equation}
\begin{split}
    \mathcal{D}_{t \to t+1}
    = \bigl(\ & (I + C_t P_{t+1}^V)^{-1} \bar{A}_t, \\
             & (I + C_t P_{t+1}^V)^{-1} (b_t - C_t\, p_{t+1}^V) \ \bigr),
\end{split}
\end{equation}
for $t=0,\dots,T-1$.
A composition
\begin{equation}
    \mathcal{D}_{i \to j} \circ \mathcal{D}_{j \to k} = \mathcal{D}_{i \to k},
\end{equation}
is defined by the composition rule
\begin{subequations}
\begin{align}
    D_{ik} &= D_{jk} D_{ij},  \\
    d_{ik} &= D_{jk} d_{ij} + d_{jk}.
\end{align}
\end{subequations}
Prefix compositions
\begin{equation}
    \mathcal{D}_{0 \to t} = \mathcal{D}_{0 \to 1} \circ \mathcal{D}_{1 \to 2} \circ \cdots \circ \mathcal{D}_{t-1 \to t},
\end{equation}
for $t=1,\cdots,T$, provides $\mathcal{D}_{0 \to t} = (D_{0t}, d_{0t})$ that allows computing $\delta x_t$ from
$\delta x_t = D_{0t}\, (\hat{x}_0 - x_0) + d_{0t}$.
Finally, $\delta \lambda_t$ can be recovered from \eqref{eq:all-state-costate}, $\delta w_t$ and $\delta \beta_t$ can be recovered from \eqref{eq:stage-delta-u-beta}, $\delta s_t$ and $\delta z_t$ can be recovered from \eqref{eq:stage-delta-s-z} and \eqref{eq:terminal-delta-s-z}.

\subsection{JAX/GPU Implementation}
We implement IP-PAS in JAX, parallelizing the primal-dual KKT step computation on the GPU. 
Additionally, our implementation follows IPOPT~\cite{ipopt}, incorporating monotone barrier-parameter updates, inertia correction, filter-based backtracking line search, second-order corrections, and a feasibility restoration phase.
These mechanisms preserve the globalization and numerical robustness of IPOPT while enabling temporal parallelization.

\section{Results and Discussion}

\subsection{Experimental Setup}
\begin{figure}[t]
    \centering

    \begin{subfigure}[b]{0.49\linewidth}
        \centering
        \includegraphics[width=0.6\linewidth]{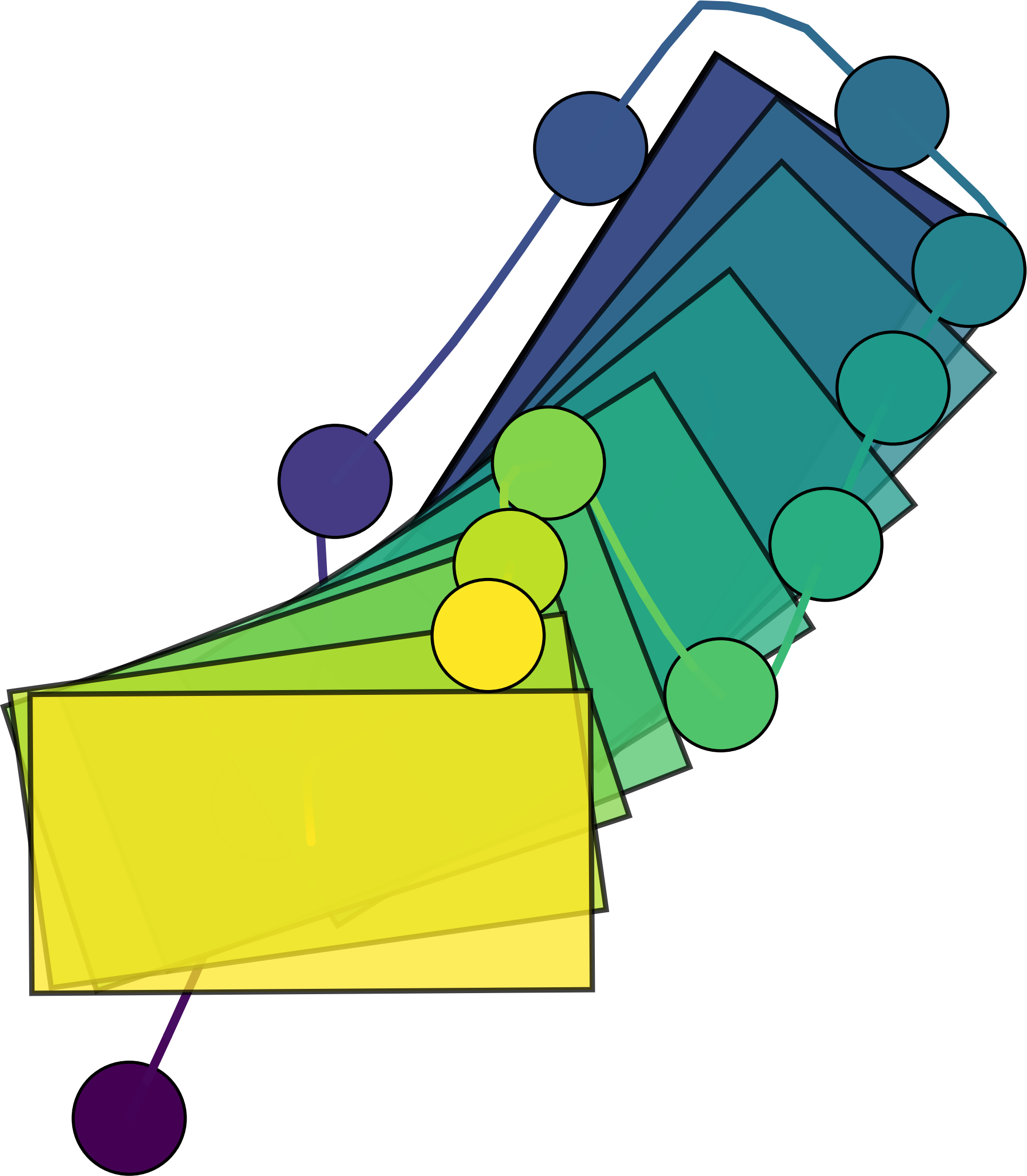}
        \caption{Push box}
        \label{fig:push_box}
    \end{subfigure}
    \hfill
    \begin{subfigure}[b]{0.49\linewidth}
        \centering
        \includegraphics[width=0.6\linewidth]{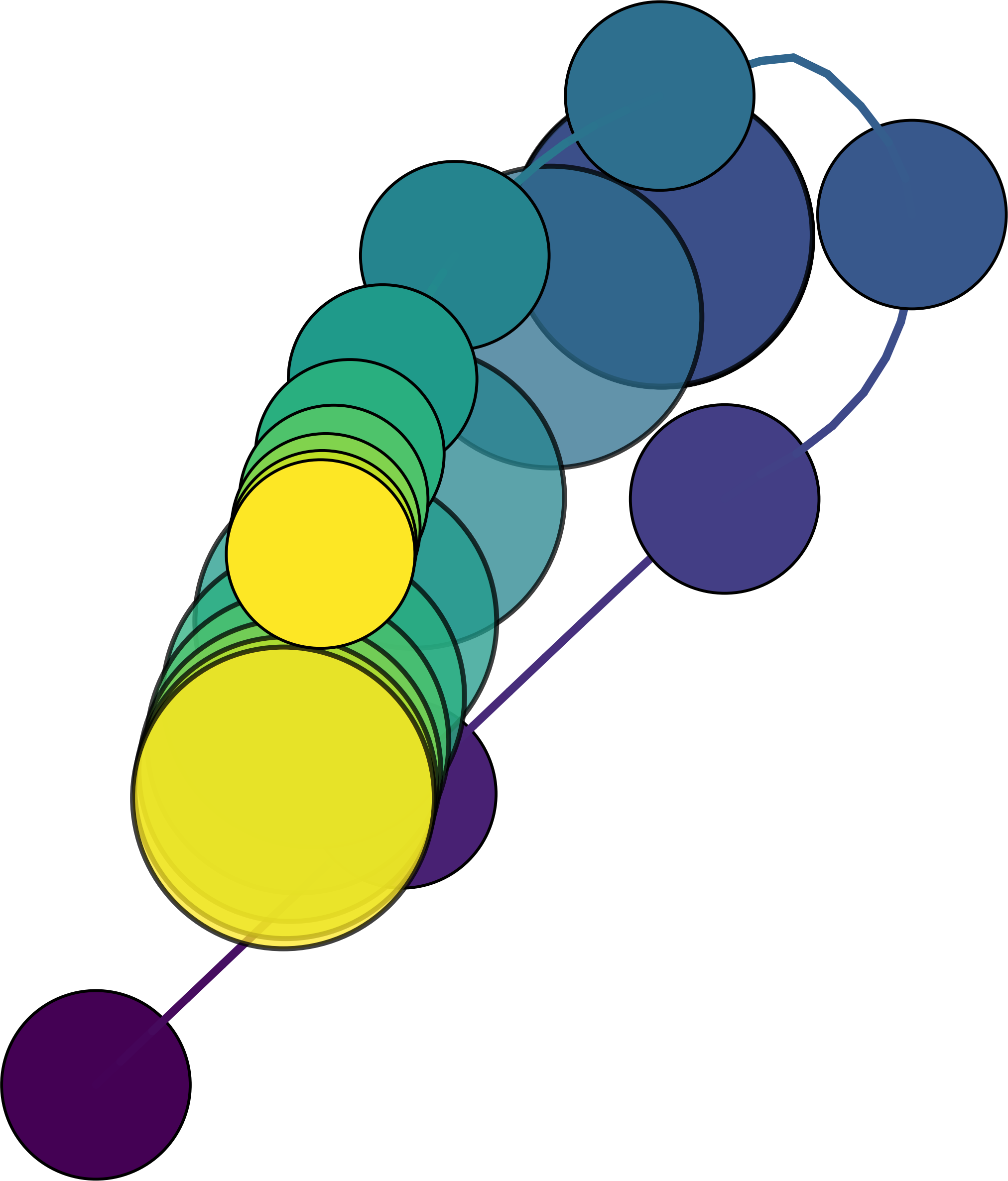}
        \caption{Push circle}
        \label{fig:push_circle}
    \end{subfigure}
    \hfill
    \begin{subfigure}[b]{0.49\linewidth}
        \centering
        \includegraphics[width=0.6\linewidth]{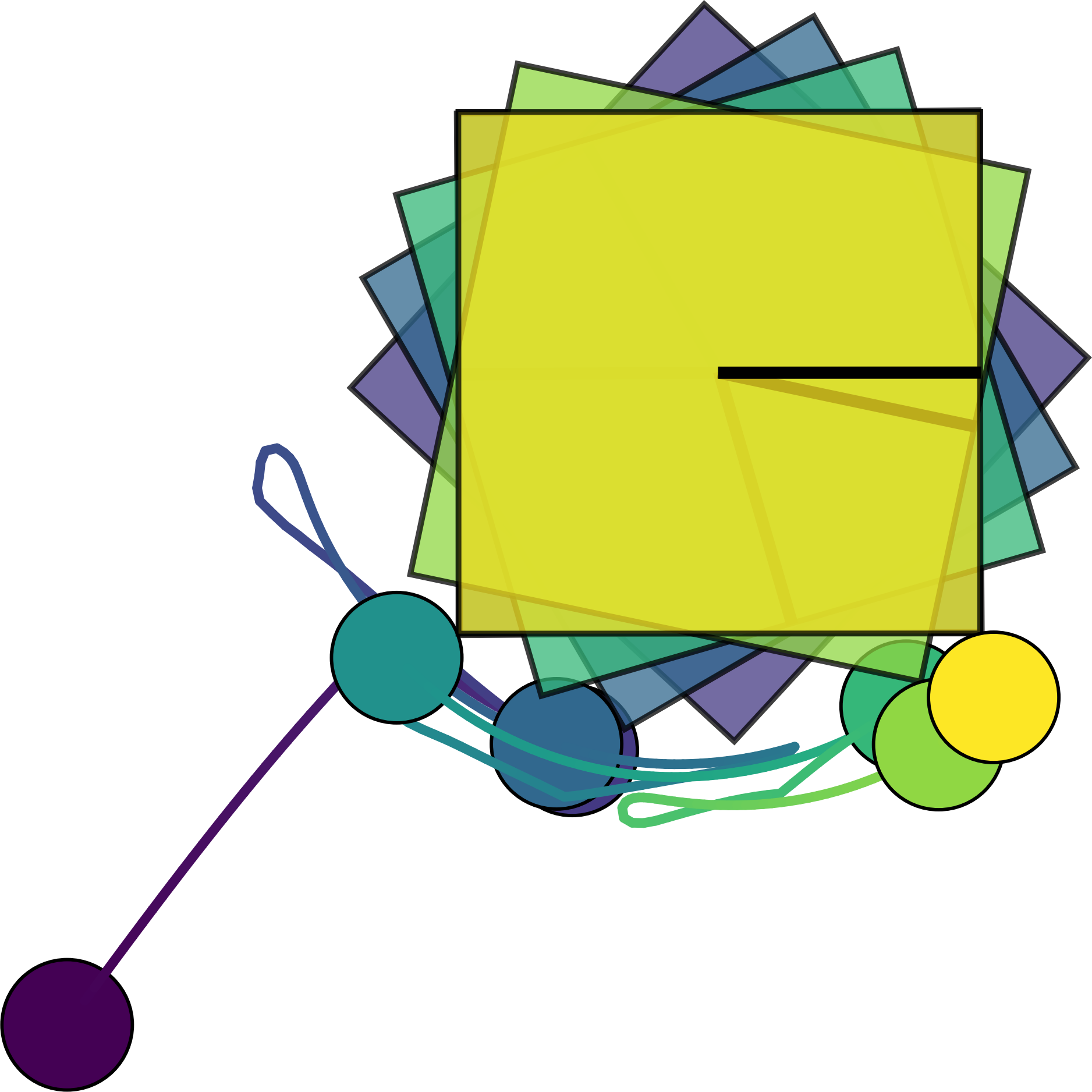}
        \caption{Square knob}
        \label{fig:square_knob}
    \end{subfigure}
    \hfill
    \begin{subfigure}[b]{0.49\linewidth}
        \centering
        \includegraphics[width=0.6\linewidth]{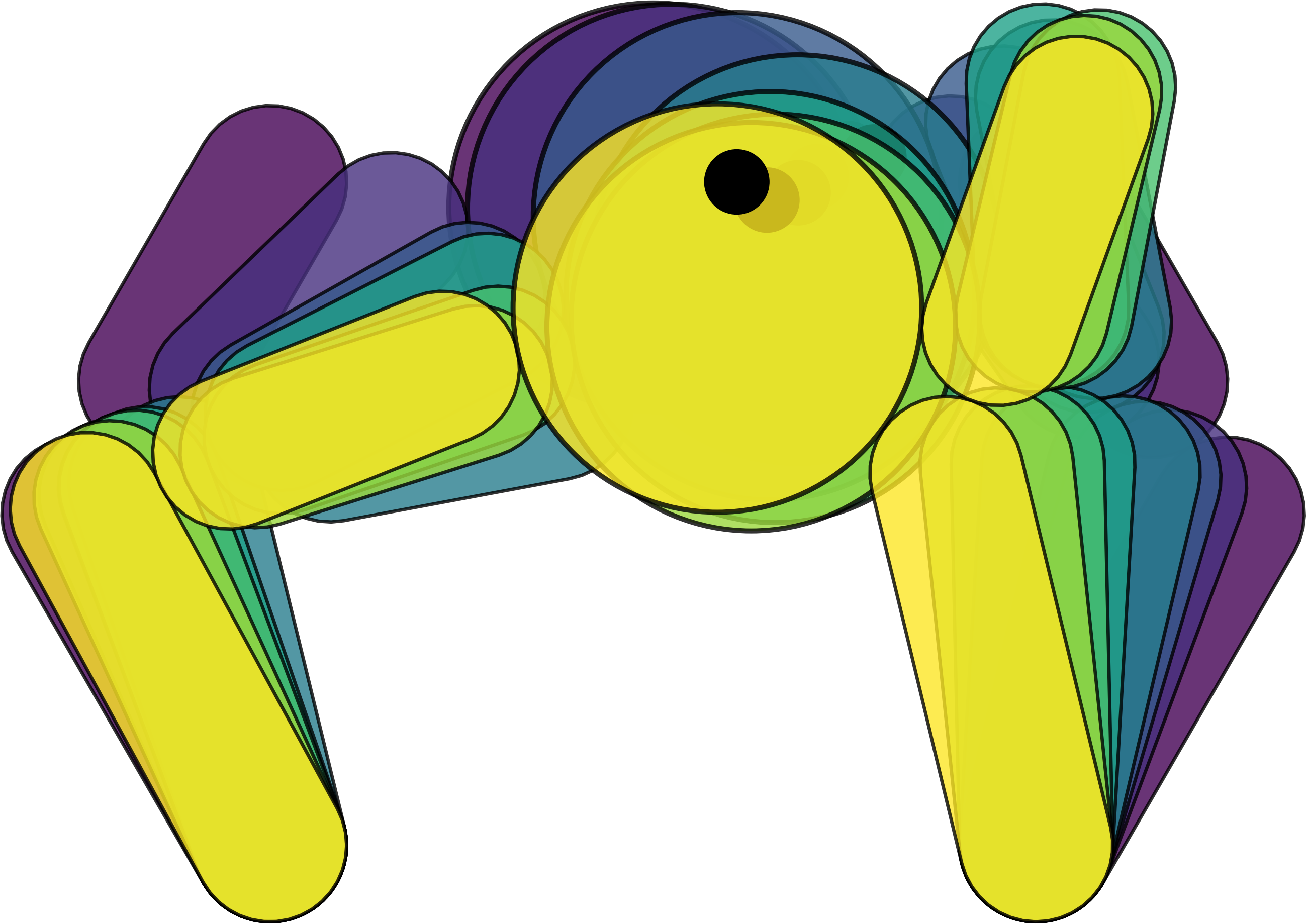}
        \caption{Planar finger}
        \label{fig:planar_finger}
    \end{subfigure}

    \caption{Planar benchmark tasks.}
    \label{fig:planar-tasks}
\end{figure}

We evaluate the proposed lifted multiple shooting transcription and IP-PAS solver on six contact-rich planning tasks: box pushing (\cref{fig:push_box}), circle pushing (\cref{fig:push_circle}), square knob manipulation (\cref{fig:square_knob}), planar finger manipulation (\cref{fig:planar_finger}), sphere in-hand manipulation (\cref{fig:inhand-tasks}a), and pen in-hand manipulation (\cref{fig:inhand-tasks}b). The in-hand manipulation benchmarks in Table~\ref{tbl:benchmark} use a 20-step horizon and are evaluated separately from the longer-horizon IP-PAS examples in Fig.~\ref{fig:inhand-tasks}. For each task and solver, we evaluate 200 randomly sampled initial conditions. All methods are initialized with a zero trajectory, without task-specific warm starts or contact-guiding costs. We measure goal error using the terminal cost $\ell_T(x_T)$ and consider a trial successful if its final goal error is below $0.01$.

We compare IP-PAS on the lifted multiple shooting transcription against single shooting, multiple shooting, and lifted multiple shooting solved using IPOPT \cite{ipopt} and \mbox{FATROP} \cite{vanroye2023fatrop}. We additionally evaluate contact-implicit formulations solved using IPOPT with Scholtes relaxation \cite{scholtes2001convergence} and IMPACT \cite{li2026impact}, which enforces \ac{MPCC} complementarity through exact analytical projection. All reported solve times include dynamics evaluations.

\begin{table}[t]
    \centering
    \caption{Benchmark results. Best and second-best values are shown in blue and green, respectively.}
    \label{tbl:benchmark}
    \setlength{\tabcolsep}{3pt}
    \resizebox{\columnwidth}{!}{%
    \begin{tabular}{cccccc}
        \toprule
        \hspace*{-5em}\makecell{Task\\(Horizon)}\hspace{-5em} & Solver & Transcription & \hspace{-5em}\makecell{Success\\rate [\%]}\hspace{-5em} & \makecell{Goal\\error} & \makecell{Solve\\time [s]} \\
        \midrule
        \multirow{8}{*}{\makecell{Push\\box\\(80)}} 
            & \multirow{4}{*}{IPOPT}
                     & Single shooting    &\ 0.0 & 0.4238 ± 0.1338 & \best{0.94 ± 0.54} \\
            &        & Multiple shooting  & 40.0 & 0.2573 ± 0.2375 & 2.30 ± 0.84 \\
            &        & Contact implicit   & 47.5 & 0.2243 ± 0.2375 & \second{1.21 ± 0.47} \\
            &        & Lifted multiple s. & \best{98.5} & \best{0.0129 ± 0.0550} & 2.43 ± 0.58 \\
            & \multirow{2}{*}{FATROP}
                     & Multiple shooting  & 33.5 & 0.2640 ± 0.2208 & 1.47 ± 0.22 \\
            &        & Lifted multiple s. & 96.0 & 0.0237 ± 0.0954 & 1.58 ± 0.26 \\
            & IMPACT & Contact implicit   & 96.0 & \second{0.0223 ± 0.0732} & 1.33 ± 0.46 \\
            & IP-PAS & Lifted multiple s. & \second{96.5} & 0.0249 ± 0.0920 & 1.43 ± 0.26 \\
        \midrule
        \multirow{8}{*}{\makecell{Push\\circle\\(100)}} 
            & \multirow{4}{*}{IPOPT}  
                     & Single shooting    &\ 0.0 & 0.1733 ± 0.0449 & 1.19 ± 0.44 \\
            &        & Multiple shooting  & 33.5 & 0.1334 ± 0.0698 & 2.25 ± 0.45 \\
            &        & Contact implicit   & 25.5 & 0.1503 ± 0.0642 & \best{0.29 ± 0.08} \\
            &        & Lifted multiple s. & 95.5 & 0.0212 ± 0.0608 & 2.64 ± 0.29 \\
            & \multirow{2}{*}{FATROP}
                     & Multiple shooting  & 34.0 & 0.1295 ± 0.0690 & 1.26 ± 0.18 \\
            &        & Lifted multiple s. & \best{100.0}\ & \second{0.0049 ± 0.0094} & 1.68 ± 0.22 \\
            & IMPACT & Contact implicit   & \best{100.0}\ & \best{0.0047 ± 0.0091} & \second{0.92 ± 0.36} \\
            & IP-PAS & Lifted multiple s. & \second{99.5} & 0.0118 ± 0.0105 & 1.23 ± 0.19 \\
        \midrule
        \multirow{8}{*}{\makecell{Square\\knob\\(400)}} 
            & \multirow{4}{*}{IPOPT}  
                     & Single shooting    &\ 0.0 & 0.7010 ± 0.8527 & 3.42 ± 1.07 \\
            &        & Multiple shooting  & 61.5 & 0.2728 ± 0.4402 & 3.92 ± 1.28 \\
            &        & Contact implicit   & 92.0 & 0.2652 ± 1.4263 & 1.69 ± 0.21 \\
            &        & Lifted multiple s. & \best{100.0}\ & \best{0.0120 ± 0.0081} & 3.49 ± 0.69 \\
            & \multirow{2}{*}{FATROP}
                     & Multiple shooting  & 67.5 & 0.2435 ± 0.4540 & 1.86 ± 0.21 \\
            &        & Lifted multiple s. & 94.0 & 0.0921 ± 0.2730 & 1.98 ± 0.51 \\
            & IMPACT & Contact implicit   & 81.5 & 0.1876 ± 0.2182 & \second{1.57 ± 0.58} \\
            & IP-PAS & Lifted multiple s. & \second{96.0} & \second{0.1033 ± 0.1168} & \best{1.22 ± 0.63} \\
        \midrule
        \multirow{8}{*}{\makecell{Planar\\finger\\(100)}} 
            & \multirow{4}{*}{IPOPT}  
                     & Single shooting    &\ 0.0 & 0.5814 ± 0.0487 & 2.33 ± 0.62 \\
            &        & Multiple shooting  & 89.5 & 0.0551 ± 0.0981 & 2.68 ± 0.79 \\
            &        & Contact implicit   & 35.0 & 0.4489 ± 0.0413 & \best{1.08 ± 0.21} \\
            &        & Lifted multiple s. & \best{99.5} & \second{0.0120 ± 0.0136} & 2.17 ± 0.45 \\
            & \multirow{2}{*}{FATROP}
                     & Multiple shooting  & 69.0 & 0.3683 ± 0.0854 & 1.67 ± 0.21 \\
            &        & Lifted multiple s. & \second{96.5} & \best{0.0105 ± 0.0219} & 2.00 ± 0.31 \\
            & IMPACT & Contact implicit   & 93.0 & 0.0463 ± 0.0602 & \second{1.18 ± 0.47} \\
            & IP-PAS & Lifted multiple s. & 91.5 & 0.0493 ± 0.0274 & 1.84 ± 0.58 \\
        \midrule
        \multirow{7}{*}{\makecell{Sphere\\in-hand\\(20)}} 
            & \multirow{4}{*}{IPOPT}  
                     & Single shooting    &\ 7.0 & 0.3155 ± 0.5281 & 8.06 ± 2.86 \\
            &        & Multiple shooting  & 13.0 & 0.2734 ± 0.5047 & 10.5 ± 3.84 \\
            &        & Contact implicit   & 14.0 & 0.5026 ± 0.4008 & \second{7.16 ± 1.06} \\
            &        & Lifted multiple s. & \best{94.5} & \best{0.0056 ± 0.0239} & 15.3 ± 3.00 \\
            & \multirow{2}{*}{FATROP}
                     & Multiple shooting  & 33.0 & 0.3472 ± 0.7692 & 9.02 ± 5.32 \\
            &        & Lifted multiple s. & 90.5 & \second{0.0211 ± 0.0579} & 12.6 ± 4.21 \\
            & IP-PAS & Lifted multiple s. & \second{92.0} & 0.0242 ± 0.0430 & \best{4.80 ± 2.29} \\
        \midrule
        \multirow{7}{*}{\makecell{Pen\\in-hand\\(20)}} 
            & \multirow{4}{*}{IPOPT}  
                     & Single shooting    &\ 0.0 & 0.2303 ± 0.9982 & 8.26 ± 2.66 \\
            &        & Multiple shooting  & 20.0 & 0.1647 ± 0.4306 & 8.37 ± 2.92 \\
            &        & Contact implicit   & 32.5 & 0.1355 ± 0.3989 & \second{7.61 ± 1.08} \\
            &        & Lifted multiple s. & \best{95.5} & \best{0.0437 ± 0.0526} & 21.6 ± 4.06 \\
            & \multirow{2}{*}{FATROP}
                     & Multiple shooting  & 22.0 & 0.1013 ± 0.3099 & 8.46 ± 2.97 \\
            &        & Lifted multiple s. & 91.0 & 0.0763 ± 0.0801 & 17.6 ± 1.04 \\
            & IP-PAS & Lifted multiple s. & \second{92.5} & \second{0.0683 ± 0.0535} & \best{5.66 ± 2.40} \\
        \bottomrule
    \end{tabular}}
\end{table}

\subsection{Comparison with Single and Multiple Shooting} 
\cref{tbl:benchmark} shows that lifted multiple shooting substantially improves the reliability of trajectory optimization through contact compared with conventional shooting formulations. With IPOPT, lifted multiple shooting achieves success rates of $98.5\%$, $95.5\%$, $100\%$, $99.5\%$, $94.5\%$, and $95.5\%$ for box pushing, circle pushing, square knob manipulation, planar finger manipulation, sphere in-hand manipulation, and pen in-hand manipulation, respectively. In comparison, conventional multiple shooting achieves success rates of only $40.0\%$, $33.5\%$, $61.5\%$, $89.5\%$, $13.0\%$, and $20.0\%$ on the same tasks. The improvement is also reflected in the final goal error, with lifted multiple shooting consistently obtaining substantially smaller errors.

The improved performance can be attributed to the explicit treatment of contact impulses as optimization variables. Rather than requiring the outer optimizer to modify the trajectory while contact impulses are determined implicitly through contact dynamics, the lifted formulation exposes the impulses directly and enforces their consistency through the equality constraint $\gamma_t=f_\text{contact}(x_t,u_t)$. This allows the optimizer to directly adjust contact impulses $\gamma_t$, which directly influence the state transition through $x_{t+1}=f_\text{integrate}(x_t,u_t,\gamma_t)$, while the contact consistency constraint $\gamma_t=f_\text{contact}(x_t,u_t)$ is progressively satisfied. Although lifting introduces additional decision variables and equality constraints, it substantially improves optimization reliability while requiring 1000$\times$ less smoothing, at only a modest increase in computational cost.

\subsection{Comparison with Contact Implicit}
Lifted multiple shooting achieves success rates comparable to those of the contact implicit baselines, particularly IMPACT, which handles \ac{MPCC} complementarity constraints through exact analytical projection. For box pushing, circle pushing, square knob manipulation, and planar finger manipulation, lifted multiple shooting with IPOPT achieves success rates of 98.5\%, 95.5\%, 100\%, and 99.5\%, respectively, compared with 96.0\%, 100\%, 81.5\%, and 93.0\% for IMPACT. IMPACT is not evaluated on the in-hand manipulation tasks because its implementation requires symbolic dynamics.

In contrast, the contact implicit transcription solved by IPOPT with Scholtes relaxation performs substantially worse, highlighting the importance of the \ac{MPCC} solution strategy for contact implicit optimization. 

Notably, lifted multiple shooting successfully handles transitions between contact modes arising from the nonsmooth geometry in box pushing and square knob manipulation. This demonstrates that explicitly enumerating or encoding individual contact faces is not always necessary. Instead, the lifted contact impulses and state-dependent contact Jacobians allow the optimizer to navigate these geometric contact-mode transitions.

IMPACT achieves shorter solve times than lifted multiple shooting with IPOPT. This is partly because the reported runtimes include both trajectory optimization solver time and dynamics evaluation time, with lifted multiple shooting incurring additional computational cost from repeated contact-resolution solves that contact implicit transcription avoids.

\subsection{IP-PAS Temporal Parallelization}

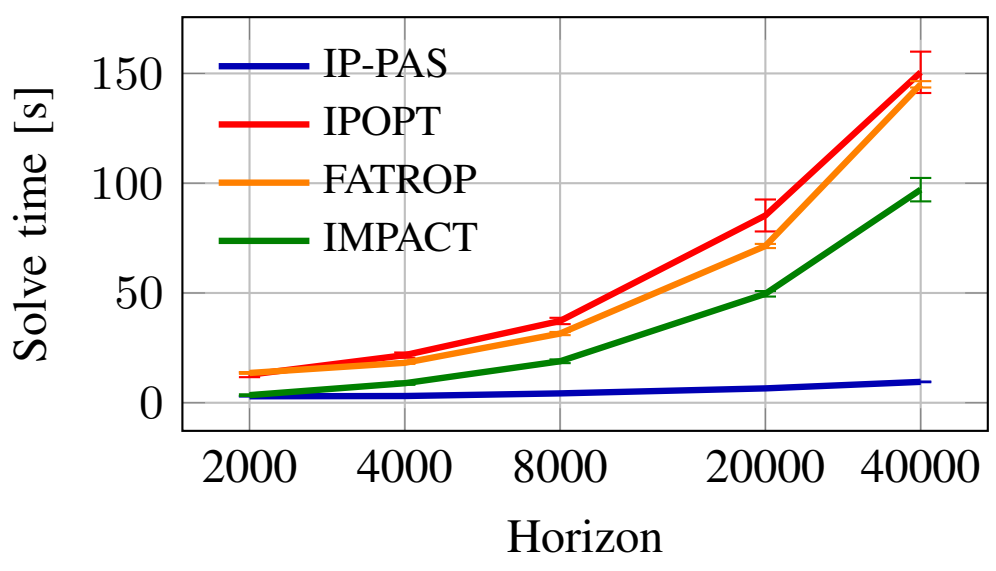
\begin{figure}[t]
    \centering
    \begin{tikzpicture}
    \begin{axis}[
        width=0.8\linewidth,
        height=0.5\linewidth,
        xlabel={Horizon},
        ylabel={Solve time [s]},
        xmode=log,
        font=\small,
        log basis x=10,
        log basis y=10,
        xtick={5,10,20,50,100},
        xticklabels={2000,4000,8000,20000,40000},
        grid=both,
        legend style={
            at={(0.02,0.98)},
            anchor=north west,
            draw=none,
            fill=none,
            font=\footnotesize,
        },
        legend cell align=left,
        every axis plot/.append style={
            line width=1.2pt,
        },
    ]

    \addplot[
        blue!70!black,
        error bars/.cd,
            y dir=both,
            y explicit,
    ] coordinates {
        (5,2.9379) +- (0,0.0388)
        (10,3.0650) +- (0,0.0313)
        (20,4.2635) +- (0,0.2309)
        (50,6.5440) +- (0,0.0417)
        (100,9.5109) +- (0,0.1624)
    };
    \addlegendentry{IP-PAS}

    \addplot[
        red,
        error bars/.cd,
            y dir=both,
            y explicit,
    ] coordinates {
        (5,12.7657) +- (0,1.1144)
        (10,21.6874) +- (0,1.1959)
        (20,37.2484) +- (0,1.4406)
        (50,85.3011) +- (0,7.2804)
        (100,150.5065) +- (0,9.4186)
    };
    \addlegendentry{IPOPT}

    \addplot[
        orange,
        error bars/.cd,
            y dir=both,
            y explicit,
    ] coordinates {
        (5,13.5635) +- (0,0.3866)
        (10,18.1887) +- (0,0.4847)
        (20,31.5120) +- (0,0.7178)
        (50,71.4024) +- (0,0.9437)
        (100,145.0371) +- (0,1.4424)
    };
    \addlegendentry{FATROP}

    \addplot[
        green!50!black,
        error bars/.cd,
            y dir=both,
            y explicit,
    ] coordinates {
        (5,3.4684) +- (0,0.3766)
        (10,8.8956) +- (0,0.7245)
        (20,18.8884) +- (0,0.8431)
        (50,49.6190) +- (0,1.2498)
        (100,97.0590) +- (0,5.3313)
    };
    \addlegendentry{IMPACT}

    \end{axis}
\end{tikzpicture}
    \caption{Solve time scaling with horizon length.}
    \label{fig:solve-time-scaling}
\end{figure}

We evaluate IP-PAS against IPOPT and FATROP for the same lifted multiple shooting transcription.
This comparison measures end-to-end solver performance, including contact-resolution and derivative evaluations, rather than isolating the temporal linear algebra alone. As shown in \cref{tbl:benchmark}, IP-PAS reduces the mean solve time relative to FATROP on all four planar tasks, with larger gains on the high-dimensional in-hand manipulation tasks.
For sphere in-hand manipulation, the mean solve time decreases from 12.6~s to 4.80~s; for pen in-hand manipulation, it decreases from 17.6~s to 5.66~s, corresponding to speedups of 2.6$\times$ and 3.1$\times$, respectively. IP-PAS maintains comparable success rates and terminal goal errors on both tasks.

We also evaluate IP-PAS on the longer-horizon in-hand sphere and pen manipulation examples in \cref{fig:inhand-tasks}, with horizons of 40 and 200 steps, respectively. IP-PAS solves these instances in 4.8~s and 7.3~s. 
To examine temporal scaling, \cref{fig:solve-time-scaling} reports solve times for square knob manipulation as the horizon increases from 2{,}000 to 40{,}000 steps. The target rotation increases with the horizon, from 5 to 100 revolutions. IPOPT and FATROP use lifted multiple shooting in this comparison, whereas IMPACT uses a contact implicit transcription. 
While the baseline's solve times increase substantially, IP-PAS exhibits much slower runtime growth. This behavior is consistent with replacing the serial backward and forward recursions with associative scans of $\mathcal{O}(\log T)$ parallel depth. 
Because the reported runtimes include the full nonlinear optimization and contact-dynamics computations, the observed gains represent practical end-to-end speedups rather than a claim of logarithmic total solve time.

\subsection{Hardware Validation}
\begin{figure}[t]
    \centering
    \includegraphics[width=0.24\linewidth]{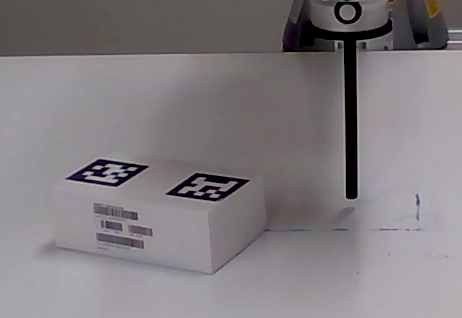}
    \includegraphics[width=0.24\linewidth]{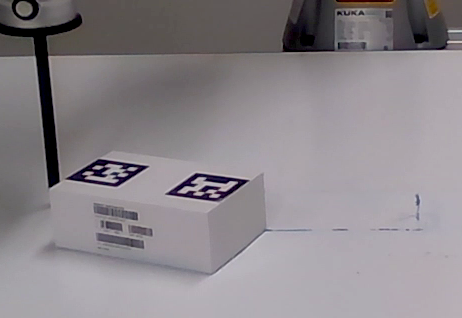}
    \includegraphics[width=0.24\linewidth]{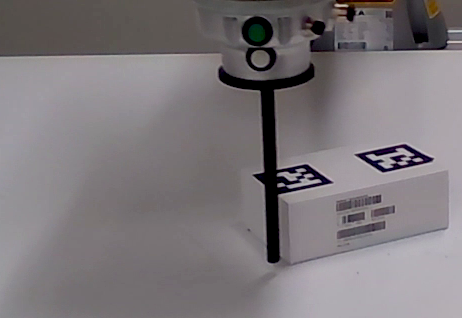}
    \includegraphics[width=0.24\linewidth]{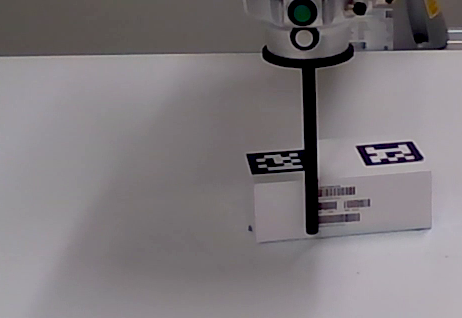}
    \caption{Hardware push box experiment.}
    \label{fig:hardware-push-box}
\end{figure}

We additionally validate the proposed approach on a physical box-pushing experiment. Figure~\ref{fig:hardware-push-box} shows the robot executing a planned pushing trajectory on the hardware, successfully moving the box from its initial configuration toward the desired target. The experiment demonstrates that the trajectories generated by the proposed formulation can be transferred to a physical system.

\section{Conclusions}

We presented a lifted multiple shooting transcription for trajectory optimization through contact together with a GPU-parallel interior-point solver based on a parallel associative scan. Across a range of contact planning and manipulation tasks, lifting contact impulses substantially improves optimization reliability over conventional single and multiple shooting transcriptions. The resulting formulation achieves success rates comparable to an MPCC-based contact-implicit solver while avoiding explicit complementarity constraints in the outer trajectory optimization problem. 
The proposed \mbox{IP-PAS} solver further preserves the success rate and solution quality of serial solvers while reducing solve time, with the largest benefit observed for high-dimensional in-hand manipulation and long-horizon tasks. Together, these results demonstrate the feasibility of explicitly exposing contact impulses and exploiting the temporal structure for trajectory optimization through contact.

\section*{Acknowledgment}
The authors thank Andrew Field for help in evaluating the multiple shooting transcription.

\end{document}